\documentclass[5p]{elsarticle}
\usepackage[utf8]{inputenc}
\usepackage[ruled,vlined]{algorithm2e}
\usepackage{bbm}
\usepackage{csvsimple}
\usepackage{amsmath}
\usepackage{graphicx}
\usepackage[export]{adjustbox}
\usepackage{hyperref}
\usepackage{verbatim}
\hypersetup{
    colorlinks=true,
    linkcolor=blue,
    filecolor=magenta,      
    urlcolor=cyan,
}

\title{A multi-agent evolutionary robotics framework to train spiking neural networks}

\author[1]{Souvik Das\corref{cor1}}
\ead{souvik@purdue.edu}
\author[2]{Anirudh Shankar}
\ead{y@purdue.edu}
\author[2]{Vaneet Aggarwal}
\ead{z@purdue.edu}

\cortext[cor1]{Corresponding author}
\address[1]{Department of Physics and Astronomy, Purdue University}
\address[2]{School of Industrial Engineering, Purdue University,  \\ West Lafayette, Indiana 47907, USA}

\begin{document}

\begin{abstract}
A novel multi-agent evolutionary robotics (ER) based framework, inspired by competitive evolutionary environments in nature, is demonstrated  for training Spiking Neural Networks (SNN). The weights of a population of SNNs along with morphological parameters of bots they control in the ER environment are treated as phenotypes. Rules of the framework select certain bots and their SNNs for reproduction and others for elimination based on their efficacy in capturing food in a competitive environment. While the bots and their SNNs are given no explicit reward to survive or reproduce via any loss function, these drives emerge implicitly as they evolve to hunt food and survive within these rules. Their efficiency in capturing food as a function of generations exhibit the evolutionary signature of punctuated equilibria. Two evolutionary inheritance algorithms on the phenotypes, Mutation and Crossover with Mutation, are demonstrated. Performances of these algorithms are compared using ensembles of 100 experiments for each algorithm. We find that Crossover with Mutation promotes 40\% faster learning in the SNN than mere Mutation with a statistically significant margin.
\end{abstract}

\maketitle

\section{Introduction}
\label{sec:Introduction}

Darwinian evolution through natural selection serves as the broad inspiration for the field of evolutionary computation in defining searches for solutions to optimization problems in high dimensional spaces. Evolutionary algorithms have been used to train the weights and biases of deep artificial neural networks~\cite{Such2017DeepLearning}. In this paper, we demonstrate the use of multi-agent evolutionary algorithms, inspired by competition in nature, to train Spiking Neural Networks (SNN) as forms of artificial intelligence. SNNs are a special class of naturally realistic ANNs that mimic the biological dynamics of discrete signaling events between neurons known as spikes \cite{Ponulak2011IntroductionApplications}. This is in contrast to currently popular ANNs which use real numbers to represent average spiking frequencies. SNNs thus allow for encoding information in the temporal sequence of spikes and offer higher computational capacity per neuron than generic ANNs. The temporal sparseness of spikes also make SNNs attractive candidates for low-energy, neuromorphic hardware implementations~\cite{Tavanaei2019DeepNetworks}. Alluring though SNNs may be, training them requires novel methods since unlike generic ANNs which use continuous and differentiable activation functions in their neurons that lend themselves to gradient descent methods for learning, SNNs define the activation mechanics of their neurons in terms of the time evolution of their membrane potentials. Hence, adapting gradient descent methods for SNNs are not trivial. This motivates us to search for nature-inspired paradigms within multi-agent Evolutionary Robotics to train them.

\subsection{Evolutionary Robotics}

The field of Evolutionary Robotics (ER) considers the co-evolution of robot morphology and intelligence within an environment of selection, inheritance and mutation~\cite{Doncieux2015EvolutionaryTo}. ER may be considered a confluence of evolutionary computation and robotics. In this work, SNNs provide the intelligence of simulated robots in a multi-agent ER arena. We demonstrate a simple ER arena, consisting of an environment and rules, where the SNNs evolve to meet the criteria for reproduction with increasing efficiency. Multi-agent arenas can be challenging to learn in since the actions of one agent affect the options available to another. This sets up indirect interaction between the agents. Our work is the first to bring SNNs to a multi-agent ER arena for effective training.

In this work, synaptic weights of the SNN and morphological parameters of the robot (henceforth referred to as the ``bot") together constitute each bot's phenotype. The phenotype is identical to the genotype in our setup. A population of initially random phenotypes are created and let loose in the ER arena as described in Section~\ref{sec:SystemModel}. We investigate two evolutionary inheritance algorithms, Mutation, and Crossover with Mutation, described in Section~\ref{sec:EvolutionaryAlgorithms}. Learning behavior is seen to emerge in a few generations, including the evolutionary signature of punctuated equilibria. This is described in Section~\ref{sec:Evaluations}. Features of the punctuated equilibria are used to compare performances of the inheritance algorithms.

\subsection{Spiking Neural Networks}

Spiking Neural Networks are considered to be the third generation of neural networks ~\cite{Maass1997NetworksModels}. The temporal sequence of spikes are known to play a role in computation in brains~\cite{Mainen1995ReliabilityNeurons, Bair1996TemporalMonkey, Herikstad2011NaturalCortex}. SNNs have found success in various pattern recognition applications, including image processing and medical diagnosis~\cite{Wysoski2010EvolvingProcessing, Gupta2007CharacterNetworks, Meftah2010SegmentationModel, Escobar2009ActionNetwork, Ghosh-Dastidar2007ImprovedDetection, Kasabov2014EvolvingStroke}. SNNs may be configured in convolutional, recurrent and deep-belief network forms as well~\cite{Tavanaei2019DeepNetworks}. SNNs are a natural fit for robotics as individual spikes can trigger discrete motor movements, and sequences of spikes at different motor neurons can articulate complex, composite motions.

Learning in SNNs is achieved by optimizing the synaptic weights and spontaneous firing rate of neurons. This may be accomplished by local methods like Spike Timing Dependent Plasticity~\cite{Hartley2006UnderstandingModel}, adaptations of gradient descent techniques~\cite{Lee2020EnablingArchitectures, Tavanaei2019DeepNetworks}, or global techniques like evolutionary algorithms~\cite{Kasabov2018Time-SpaceNeurosystems}. Gradient descent techniques rely on differentiable surrogates for the SNN activation mechanism~\cite{Bohte2000Error-BackpropagationNeurons, Ponulak2010SupervisedShifting., Mohemmed2012Span:Patterns}. Although surrogate gradients have paved the way to perform training, the problem of training multi-layered SNNs efficiently remains challenging. While some forms of evolutionary algorithms have been used to train SNNs, our work distinguishes itself by the use of a multi-agent ER framework.

\subsection{Main Contributions}

The main contributions of this work are as follows.

\begin{enumerate}

\item Demonstration of a multi-agent ER framework, inspired by competition in nature, to train SNNs. The framework is kept as simple as possible with the smallest set of parameters so we may arrive at general conclusions.

\item Quantitative characterization of evolutionary learning by fitting punctuated equilibria to logistic curves.

\item Comparison between the performances of two evolutionary algorithms for training the SNNs: Mutation versus Crossover with Mutation.

\end{enumerate}

\section{System Components}
\label{sec:SystemModel}

The multi-agent ER framework within which we investigate the efficacy of evolutionary algorithms for SNN training is described in this section. Experiments are performed in a simulated arena consisting of a group of bots, each with an SNN, competing for the capture of ``food" in a ``game environment" with certain rules. The bot is described in Section~\ref{sec:Bots}, the SNN is described in Section~\ref{sec:SNN}, and the game environment and food in Section~\ref{sec:Environment}. Since the food is replenished after a capture event, the experiments can run indefinitely. The rules of evolution that kick in at each capture event are described in Section~\ref{sec:EvolutionaryAlgorithms}.

\subsection{Bots}
\label{sec:Bots}

\begin{figure}[tbph]
\centering
\includegraphics[width=\linewidth, frame]{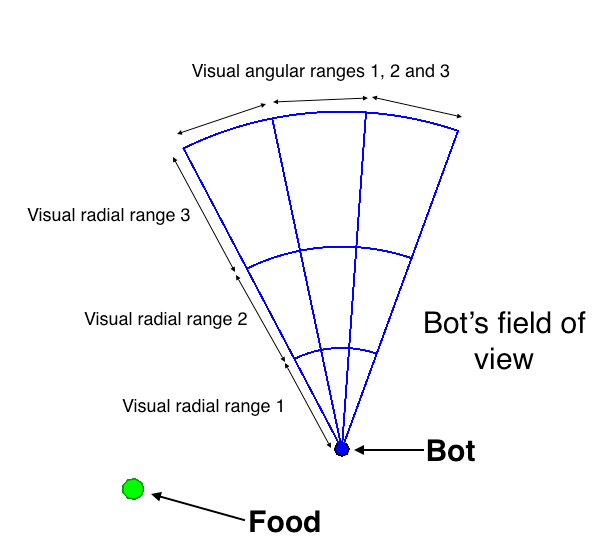}
\caption{Graphical representation of a bot. The circular dot represents its areal extent in the game environment. The blue quadrant represents its field of view that is segmented as described in Section~\ref{sec:Bots}. Each visual segment activates a different combination of the sensory neurons of its SNN.}
\label{fig:Bot}
\end{figure}

Each bot occupies a circular area (of 40 units) and has a position $(x, y)$ and angular orientation $\theta$ within a 2D game environment (of 500 units $\times$ 500 units). The movement of the bot, in response to sensory input, is governed by motor output from the SNN that controls it. Its sensory input is received through its field of view as illustrated in Fig.~\ref{fig:Bot}. The field of view is segmented into 9 parts; 3 radial ranges, and 3 angular ranges. The 3 radial ranges extend from 0 - 30, 30 - 60, and 60 - 100 units. The presence of food within the field of view triggers a different neuron for each of the radial ranges. The opening angle of the field of view, $v$, is considered a morphological parameter of the bot and is allowed to evolve along with its SNN. The angle is trisected for 3 angular ranges and the presence of food within each of them triggers a different sensory neuron. Thus, a total of 6 sensory neurons are dedicated for the bot's vision.

Four motor neurons control the movement of the bot. The first one, when fired, advances the bot by 1 unit in its orientation direction. The second makes the bot take 1 step back. The third and the fourth rotate the bot clockwise and anti-clockwise by 0.1 radians, respectively.

\subsection{The Spiking Neural Network}
\label{sec:SNN}

\begin{figure}[tbph]
\centering
\includegraphics[width=\linewidth]{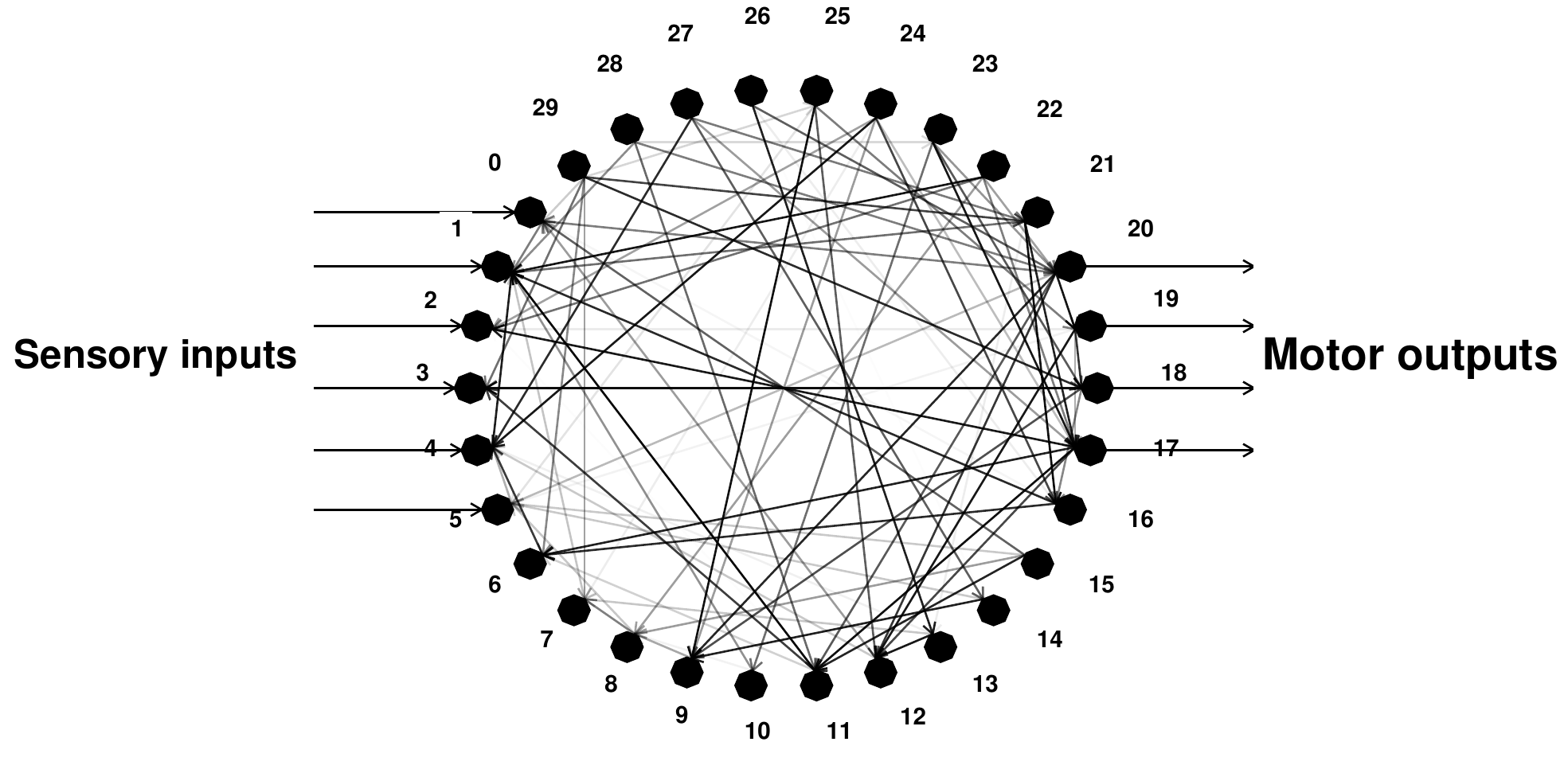}
\caption{The structure of a SNN that controls a single bot shown at a representative state in its evolution. It consists of 30 spiking neurons in a directed network. The shade of the edges correspond to their weights at a particular generation. 6 neurons are connected to the sensory inputs of the bot and 4 to its motor output.}
\label{fig:SNN}
\end{figure}

\begin{figure}[tbph]
\centering
\includegraphics[width=\linewidth]{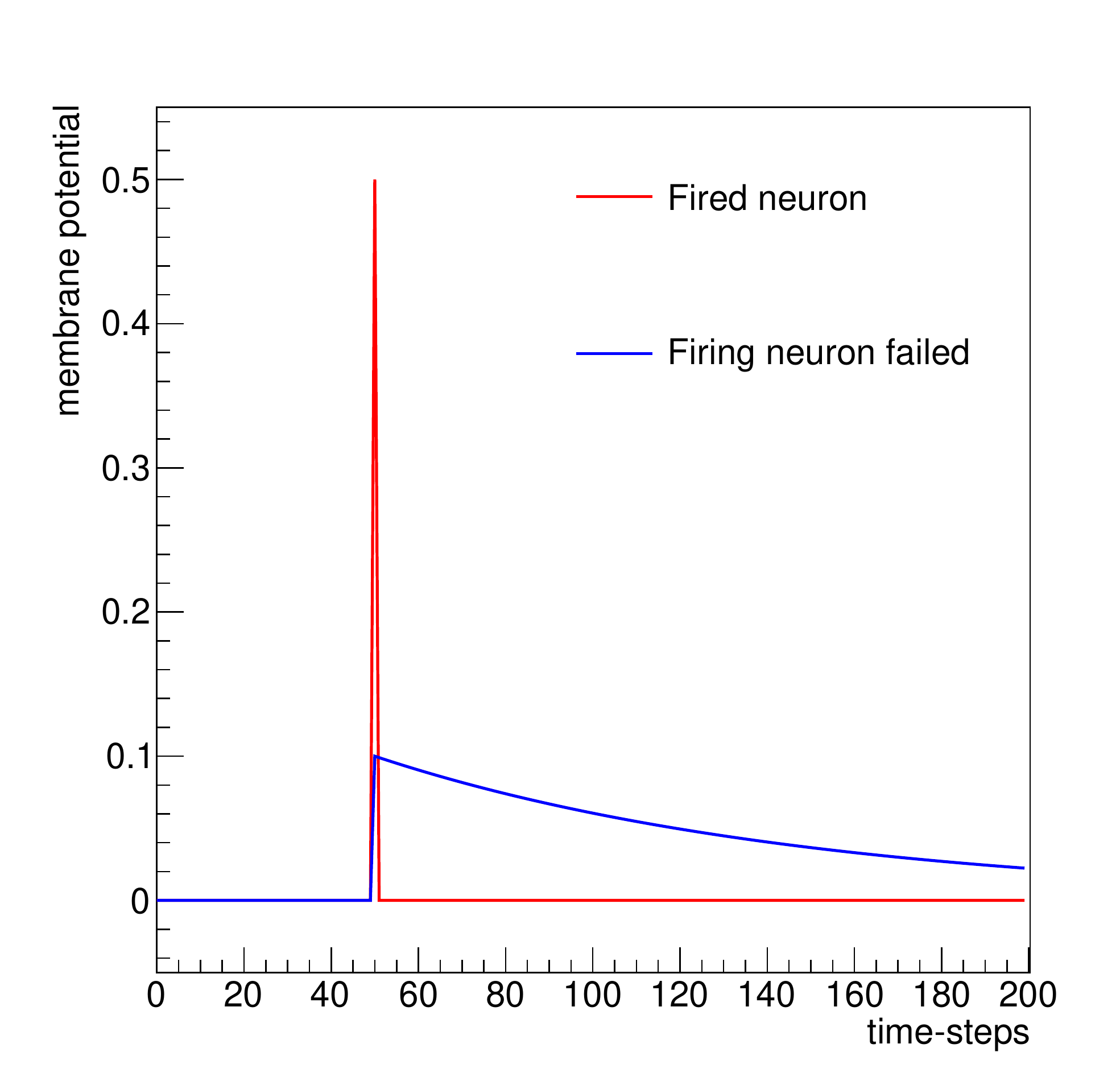}
\caption{The membrane potential of a single simulated neuron as a function of time-steps when fired with two values of incoming charge. In red is when the incoming charge corresponds to an increase in potential that exceeds the threshold $V_{th}$, and in blue is when it does not.}
\label{fig:spiking_neuron_potential}
\end{figure}

Each bot has a SNN that controls it. Each SNN consists of 30 neurons in a fully-connected, directed network, as illustrated in Fig.~\ref{fig:SNN}. The edges of the network are associated with weights $w_{ij}$, and this matrix is allowed to evolve. The network is not recurrent, hence $w_{ii} = 0$. Of the neurons, 6 are sensory and 4 are motor as has been described. The SNN operates in discrete time steps that also correspond to time steps in the motion of the bot. Each neuron has a membrane potential $V(t)$ whose dynamics is governed by the Leaky Integrate and Fire (LIF) model~\cite{Kasabov2018Time-SpaceNeurosystems}. The LIF model may be described by
\begin{equation}
\frac{dV(t)}{dt} = \frac{1}{C_m}\frac{dq}{dt} - \frac{V(t)}{R_m C_m}
\label{lif_equation}
\end{equation}    
where $dq/dt$ is the input current, $C_m$ is a measure of the neuron's membrane capacitance, and $R_m$ is its membrane resistance. The first term expresses the increase in membrane potential from the rate of charge deposition from incoming spikes. The second term reflects the decay of membrane potential due to the spontaneous neutralization of charge. In our model, we approximate LIF in the limit of infinitesimal time-steps using the difference equation:
\begin{equation}
V(t+1) - V(t) = q(t) - \beta V(t)
\label{lif_approx}
\end{equation}
where $\beta$ contains the $R_m C_m$ decay constant and is set to 1\%. The capacitance is set to unity in our simulation with no loss of generality as the scale of $V$ is set by the voltage threshold $V_{th}$ beyond which the neuron fires.

The incoming charge for neuron $i$ at time-step $t$ is given by the the sum of arriving spikes weighted by $w_{ij}$
\begin{equation}
q_i(t) = \sum_{j} w_{ij} A_j(t)
\label{lif_membrane_potential}
\end{equation}
where $A_j(t)$ is 1 if the $j^{th}$ neuron has fired in time-step $t$ and 0 otherwise.

A neuron fires if its membrane voltage exceeds the threshold $V_{th} = 0.4$ or randomly at a spontaneous rate of $b \approx 1\%$. The spontaneous rate, which corresponds loosely to the bias term for each neuron in traditional neural networks, is found to be important to avoid trapping the SNN in states where no neurons are firing or where all neurons are firing. This spontaneous firing rate, $b$, is allowed to evolve. Thus, for the $i^{th}$ neuron at time $t$,
\begin{equation}
A_i(t) = 
\begin{cases}
 1 & \text{if $V_i(t) > V_{th}$ OR $r > b$} \\
 0 & \text{otherwise} \\
\end{cases}
\label{delta_function}
\end{equation}
where $r$ is a uniform random number from 0 to 1. When the neuron fires, the membrane potential $V$ is set back to 0 at the next time-step. We illustrate the firing behavior of a single simulated neuron in Fig.~\ref{fig:spiking_neuron_potential} by plotting its membrane potential by time-step when fired with an incoming charge corresponding to potential increases greater and lesser than $V_{th}$.

\subsection{The Environment}
\label{sec:Environment}

\begin{figure}[tbph]
\centering
\includegraphics[width=\linewidth, frame]{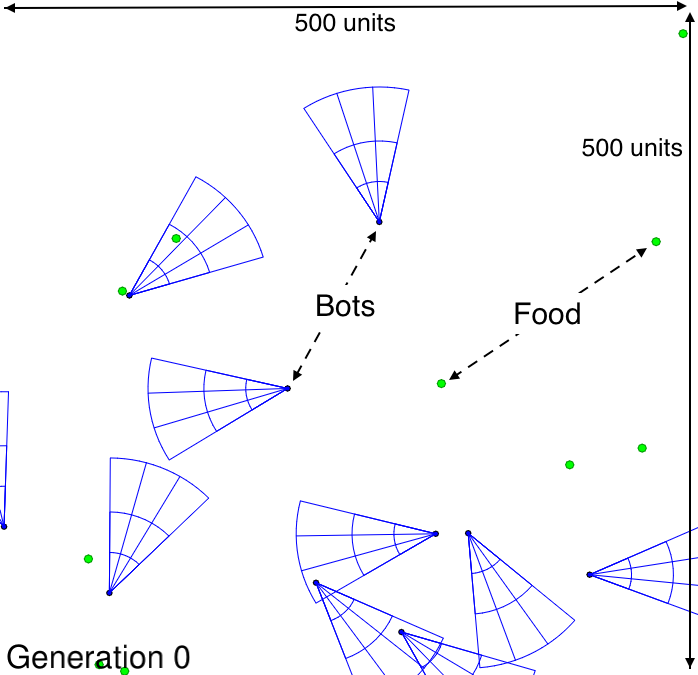}
\caption{Snapshot of the multi-agent environment within which the bots and their SNNs evolve. The physical space is 500 units x 500 units, and is populated here with 10 bots and 10 pieces of food, all in constant motion. The walls are reflective, as described in the text. The rules of evolution implemented by the environment are described in Section~\ref{sec:EvolutionaryAlgorithms}.}
\label{fig:Environment}
\end{figure}

The evolutionary environment in which our bots operate is a 2D square of 500 units $\times$ 500 units, as shown in Fig.~\ref{fig:Environment}. The walls are reflective, i.e. when bots run into the vertical walls their $\theta$ is changed to $\pi - \theta$, and when they run into horizontal walls their $\theta$ is multiplied by -1.

The environment contains entities that result in the reproduction of a bot if captured. We call these entities ``food" for the remainder of the paper. Like the bots, they each have $(x, y)$ coordinates, a fixed orientation angle $\theta$, and a randomly chosen speed. A capture occurs when the square of the Pythagorean distance between a food and a bot, $(x_{bot} - x_{food})^2 + (y_{bot} - y_{food})^2$, is less than 13. The procedures implemented in the reproduction of the bots at each capture is described in Section~\ref{sec:EvolutionaryAlgorithms}. The food is replenished in the environment by placing a new instance in a random position and orientation with a randomly chosen speed.

\section{Evolutionary Algorithms}
\label{sec:EvolutionaryAlgorithms}

Two evolutionary inheritance algorithms are investigated in this paper: we call the first one ``Mutation" and the second one ``Crossover with Mutation". In both algorithms, when a capture event occurs as described in Section~\ref{sec:Environment}, three procedures kick in: a selection procedure, a reproduction procedure, and an elimination procedure. This results in a new ``generation" of bots which then continue to compete in the environment till the next capture event. The phenotypical parameters of the bots, i.e., its SNN weight matrix $w$, the spontaneous firing rate $b$, and its visual angle $v$, are initially random. Thus, initially, there is there no correlation between what a bot senses in its field of view and what it does; its movements are random. \textit{No explicit reward} is given to the bots when it captures food. The successful bot(s) are reproduced with purely random mutations, depending on the inheritance algorithm, and bot(s) eliminated according to a fitness function to keep the population constant. With this bare minimum of evolutionary pressure, we expect the SNNs to learn to drive the bots to food with increasing efficiency in the course of a few generations. The fitness function used is
\begin{equation}
\label{eq:Fitness}
f = N / \tau,
\end{equation}
where $N$ is the number of times it has captured food and $\tau$ is its age in time-steps. During elimination, bots with the lowest values of $f$ are removed from memory.

\subsection{Mutation}
\label{sec:Mutation}

In this inheritance algorithm, the bot that captured food is selected for reproduction. The phenotype of the bot is duplicated with random mutations to create a new bot. Components of the weight matrix $w$ and $b$ are modified with random Gaussian variations of standard deviation $\mu_{mod}$. The visual angle, $v$, is modified similarly with the parameter $\mu_{visual}$. This is summarized in Algorithm~\ref{algo:Mutation}. One bot in the population is removed according the fitness function $f$ as described earlier.

\begin{algorithm}[tbph]
\SetAlgoLined
{\bf Input: } Selected bot has phenotype $(w^{old}, b^{old}, v^{old})$\;
{\bf Output: } New bot made with phenotype $(w^{new}, b^{new}, v^{new})$\;
\For{each connection (i, j) in $w$} {
  $w^{new}_{ij} \gets w^{old}_{ij} + \mathcal{N}(0, \mu_{mod})$\;
}
$b^{new} \gets b^{old} + \mathcal{N}(0, \mu_{mod})$\;
$v^{new} \gets v^{old} + \mathcal{N}(0, \mu_{visual})$\;
\caption{Evolutionary inheritance algorithm of Mutation}
\label{algo:Mutation}
\end{algorithm}

\subsection{Crossover with Mutation}
\label{sec:CrossoverMutation}

\begin{figure}[tbph]
\centering
\includegraphics[width=\linewidth]{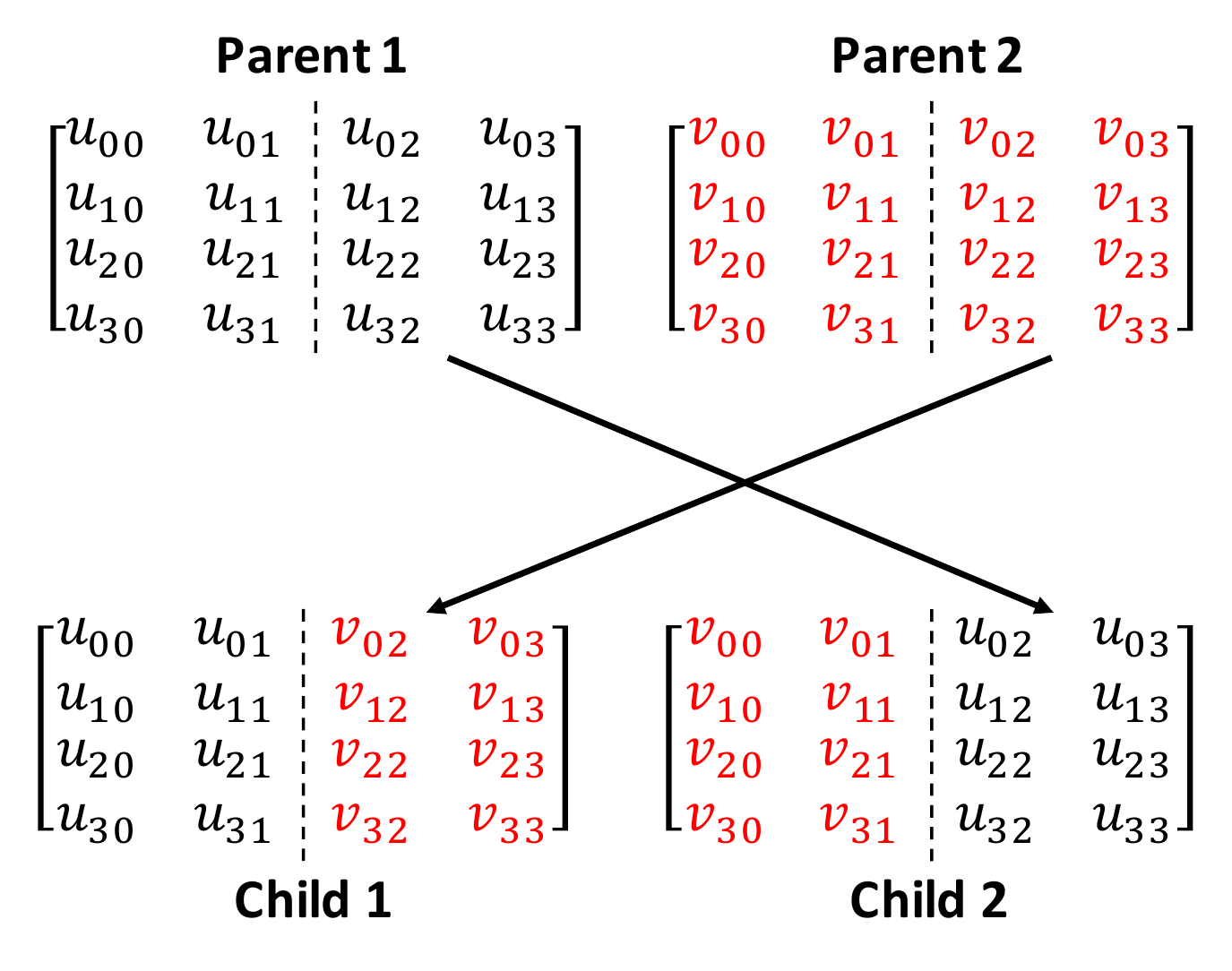}
\caption{Illustration of the crossover procedure that mixes the SNN weights of two bots for the ``Crossover and Mutation" strategy described in Section~\ref{sec:CrossoverMutation}. While the illustration is with $4 \times 4$ matrices, the SNN weight matrices are $30 \times 30$.}
\label{fig:crossover_type1}
\end{figure}

This inheritance algorithm involves waiting for two bots to capture food and mixing their phenotype parameters to create two child bots. Two bots with the lowest fitness values, $f$ as described in Eq.~\ref{eq:Fitness}, are then eliminated. There are exists a variety of crossover operations on matrices that exist in literature \cite{Tsai2015AProblem}. In our algorithm, the weight matrices of the SNNs of the two bots, $w^1$ and $w^2$, are partitioned in half and interchanged as illustrated in Fig.~\ref{fig:crossover_type1}. The new weight matrices, spontaneous rate and visual angle are then mutated in exactly the same way and with the same parameters $\mu_{mod}$ and $\mu_{visual}$ as described in Section~\ref{sec:Mutation}. This is summarized in Algorithm~\ref{algo:Crossover}.

\begin{algorithm}[tbph]
\SetAlgoLined
{\bf Input: } Selected bots have phenotypes $(w^1, b^1, v^1)$, $(w^2, b^2, v^2)$\;
{\bf Output: } New bots made with phenotypes $(w^3, b^3, v^3)$, $(w^4, b^4, v^4)$\;
$w^3 \gets w^1$\;
$w^4 \gets w^2$\;
\For{each connection (i,j) in $w^1$}{
  \If {$j > k/2$} {
    $w^4_{ij} \gets w^1_{ij}$\;
    $w^3_{ij} \gets w^2_{ij}$\;
  }
  $w^3_{ij} \gets w^3_{ij} + \mathcal{N}(0, \mu_{mod})$\;
  $w^4_{ij} \gets w^4_{ij} + \mathcal{N}(0, \mu_{mod})$\;
}
$b^3 \gets b^1 + \mathcal{N}(0, \mu_{mod})$\;
$b^4 \gets b^2 + \mathcal{N}(0, \mu_{mod})$\;
$v^3 \gets v^1 + \mathcal{N}(0, \mu_{visual})$\;
$v^4 \gets v^2 + \mathcal{N}(0, \mu_{visual})$\;
\caption{Evolutionary inheritance algorithm of Crossover with Mutation.}
\label{algo:Crossover}
\end{algorithm}
\section{Evaluations}
\label{sec:Evaluations}

We evaluate the two evolutionary algorithms by measuring the average number of time-steps, $T$, needed by a bot to capture food at each generation. For our analysis, since Crossover with Mutation requires 2 bots to capture food to advance a generation, we consider the time taken for 2 consecutive captures in the definition of $T$ for both strategies. Thus, we define $T$ as
\begin{equation}
\label{eq:T}
T = \langle t_2 - t_1 \rangle_{50\ \mathrm{generations}},
\end{equation}
where $t_1$ is the time-step at which piece of food is captured by a bot, and $t_2$ is the time-step at which another piece of food has been captured by any other bot and then yet another piece captured by any bot. This quantity is averaged over 50 generations and studied. As the SNNs learn, this is expected to decrease with the number of generations. Since this is evolutionary learning, we also expect features of punctuated equilibria which we fit to the logistic function.

Experiments for this paper are conducted in the previously described 500 units $\times$ 500 units environment with 10 bots and 5 pieces of food. Each bot is controlled by a SNN. The global mutation parameters, $\mu_{mod}$ and $\mu_{visual}$ defined in Section~\ref{sec:Mutation}, are set to 0.05 and 0.008, respectively. We arrived at these values by rough optimization of the final $T$ after 10,000 generations of evolution to obtain a fairly efficient learning environment. Our results, especially in their qualitative features, do not lose generality in the neighborhood of this parameter set.

\subsection{One experiment of Mutation}

\begin{figure}[tbph]
\centering
\includegraphics[width=\linewidth]{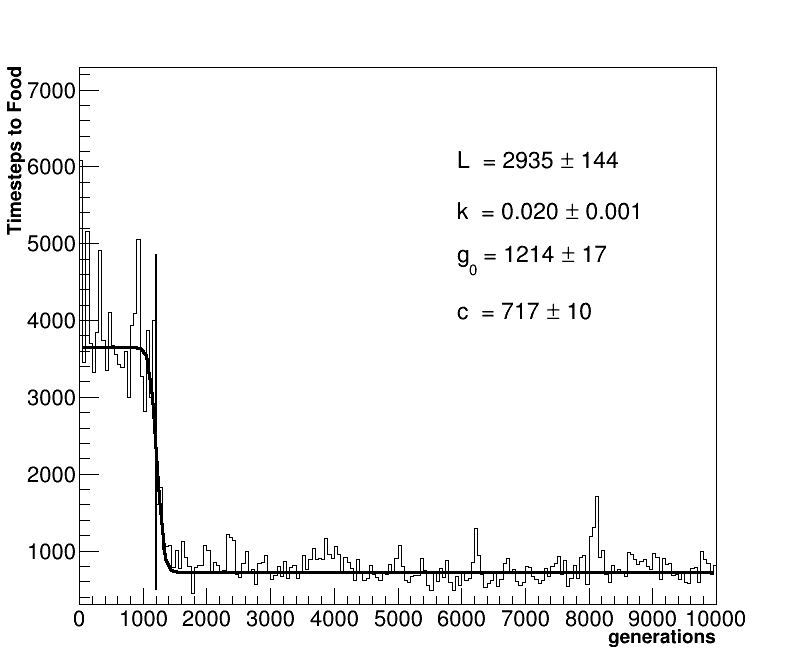}
\caption{The average time-steps to capture food, $T$ as defined in Eq.~\ref{eq:T}, as a function of the number of generations using the evolutionary inheritance algorithm of Mutation. A fit to a logistic function is used to extract quantitative features of the punctuated equilibria.}
\label{fig:training_mutation}
\end{figure}

Much can be learned by observing the outcome of one experiment with the Mutation inheritance algorithm. As seen in the \href{https://youtu.be/JCUczJRtb0I}{\textbf{video}} accompanying this paper, bots are initially seen to execute random motions with no regard for food in their fields of view. As bots accidentally capture food and reproduction begins, small mutations in a bot's phenotype that make food capture more probable allow that bot to have more offspring. Thus, after roughly 100 generations, food capture becomes less accidental and more apparently intentional as the SNNs structure themselves to make use of sensory data from their field of view. Around generation 1,400, we observe the development of hunting behavior as the bots learn to cover ground and spin their fields of view in search of food. Bots that do not hunt have lower fitness values $f$ and are eventually culled. This development results in a rapid improvement in efficiency and decrease in $T$, as defined in Eq.~\ref{eq:T}.

In Fig.~\ref{fig:training_mutation}, we study the variation of $T$ as a function of generation up to 10,000 generations. While there is a large variance in $T$ initially, as the population of bots branches into lineages that sometimes work well and sometimes do not, we note a sharp drop around generation 1,214 when a bot discovers hunting. Thereafter, the bot that discovered hunting dominates the population with its offspring and $T$ remains relatively stable up to 10,000 generations. Thus, we observe two periods of equilibrium connected by a punctuation, as expected in evolutionary systems. We extract broad features of this punctuated equilibria by fitting the graph with a logistic function on a flat pedestal of the form 
\begin{equation}
\label{eq:Logistic}
f(g) = \frac{L}{1 + e^{k(g - g_0)}} + c.
\end{equation}
The center of the punctuation, or the Inflection Point, is given by $g_0$ in generations. The sharpness of the punctuation is given by the slope of the inflection, $k$. The final equilibrium value of $T$ is given by $c$, and we call this the Convergence Point. The initial equilibrium value of $T$ is given by $L + c$. A minimum $\chi^2$ fit returns $g_0 = 1214 \pm 17$ generations, $k = 0.020 \pm 0.001$ time-steps / generation, $L = 2935 \pm 144$ time-steps and $c = 717 \pm 10$ time-steps.

\subsection{One experiment of Crossover with Mutation}

\begin{figure}[tbph]
\centering
\includegraphics[width=\linewidth]{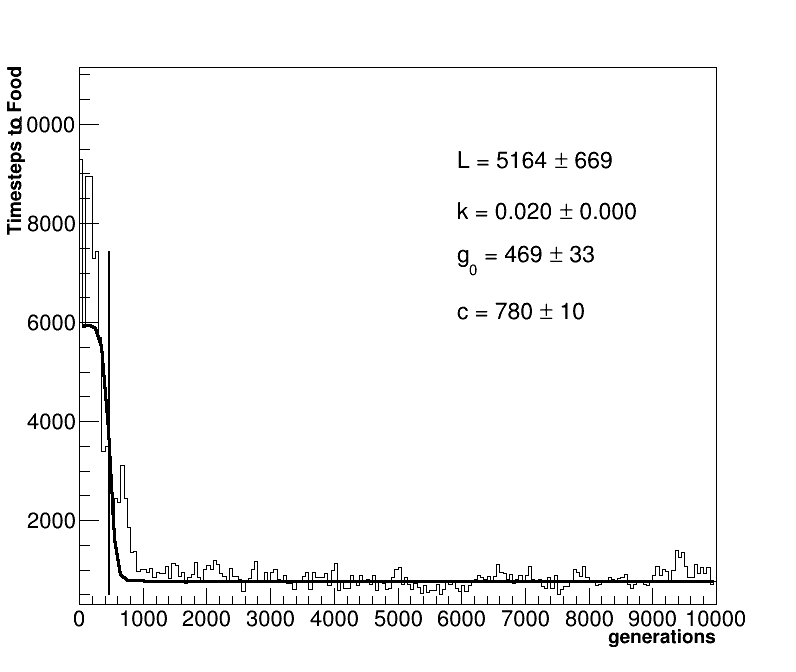}
\caption{The average time-steps to capture food, $T$, as a function of the number of generations using the evolutionary strategy of Crossover and Mutation. A fit to a logistic function is used to extract quantitative features of the punctuated equilibria.}
\label{fig:training_mutation_and_crossover}
\end{figure}

We repeat the experiment with the inheritance algorithm of Crossover with Mutation and observe similar behavior in the bots as they learn to capture food. Faster learning is observed as hunting behavior emerges around generation 469. A minimum $\chi^2$ fit with Eq.~\ref{eq:Logistic} returns $g_0 = 469 \pm 33$ generations, $k = 0.020 \pm 0.000$ time-steps / generation, $L = 5164 \pm 669$ time-steps and $c = 780 \pm 10$ time-steps.

\subsection{Comparison over experimental ensembles}

\begin{figure}[tbph]
\centering
\includegraphics[width=\linewidth]{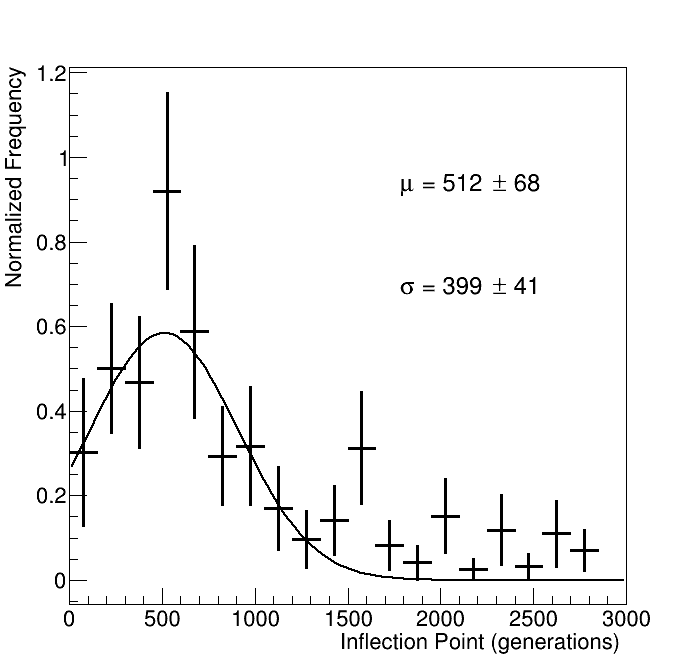}
\caption{Distribution of Inflection Points (in generations) in an ensemble of 100 experiments with the evolutionary strategy of Mutation. The histogram is binned by 150 generations and fitted with a Gaussian to estimate its mean and standard deviation.}
\label{fig:inflection_point_mutation}
\end{figure}

\begin{figure}[tbph]
\centering
\includegraphics[width=\linewidth]{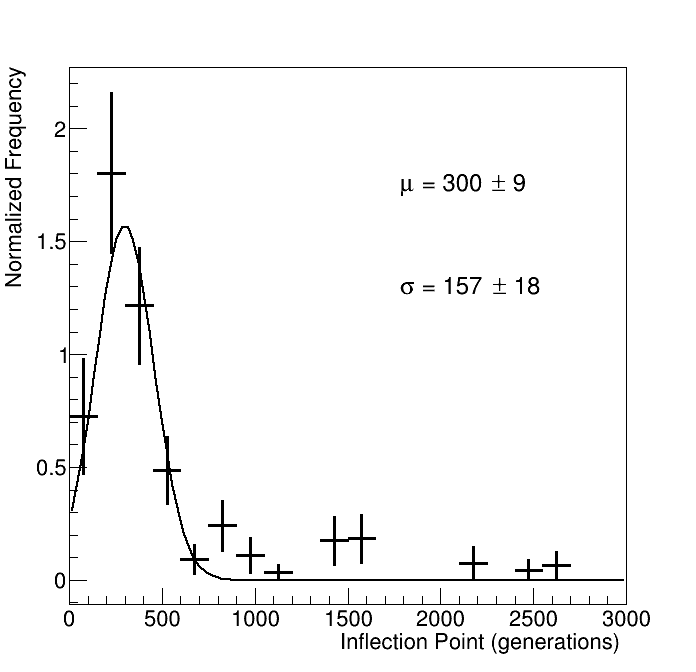}
\caption{Distribution of Inflection Points (in generations) in an ensemble of 100 experiments with the evolutionary strategy of Crossover and Mutation. The histogram is binned and fitted identically to Fig.~\ref{fig:inflection_point_mutation}}
\label{fig:inflection_point_mutation_and_crossover}
\end{figure}

\begin{figure}[tbph]
\centering
\includegraphics[width=\linewidth]{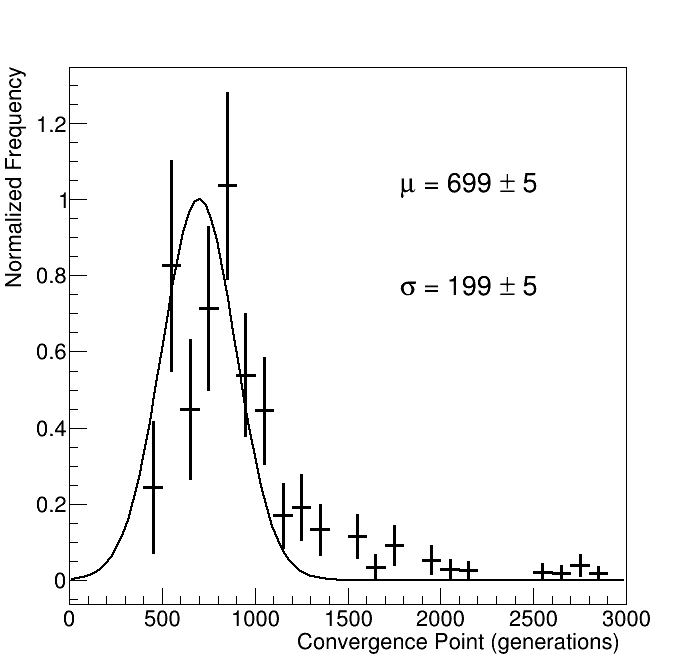}
\caption{Distribution of Convergence Points (in time-steps) in an ensemble of 100 experiments with the evolutionary strategy of Mutation. The histogram is binned by 100 time-steps and fitted with a Gaussian to estimate its mean and standard deviation.}
\label{fig:c_convergencePoint_mutation}
\end{figure}

\begin{figure}[tbph]
\centering
\includegraphics[width=\linewidth]{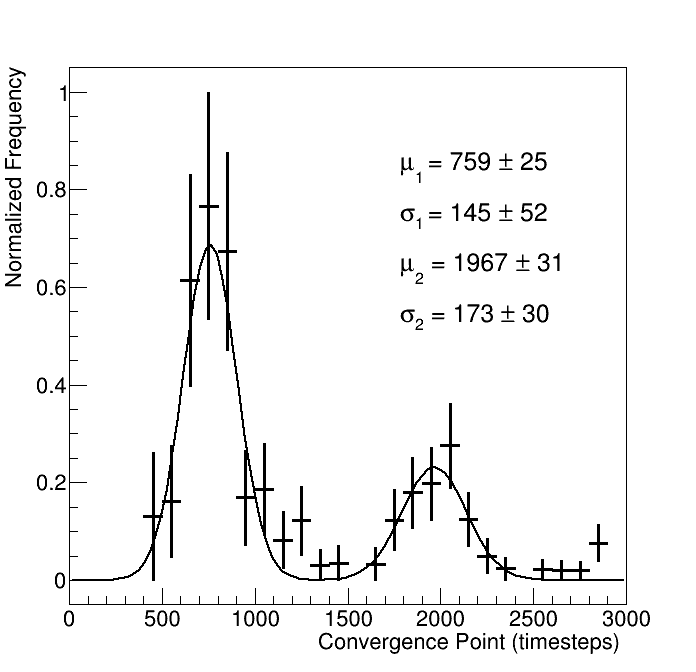}
\caption{Distribution of Convergence Points (in time-steps) in an ensemble of 100 experiments with the evolutionary strategy of Crossover and Mutation. The histogram is binned by 100 time-steps. It is bi-modal and fitted with two Gaussians. Their means and standard deviations are reported.}
\label{fig:c_convergencePoint_MutationCrossover}
\end{figure}

\begin{table*}[tbph]
\centering
\begin{tabular}{c c c}
\hline \hline
Evolutionary Strategy & Inflection Point & Convergence Point \\
& (generations)    & (time-steps) \\
\hline
Mutation &  $512 \pm 68$ & $699 \pm 5$ \\
Crossover with Mutation & $300 \pm 9$  & $759 \pm 25$ and $1967 \pm 31$ \\
\hline \hline
\end{tabular}
\caption{Summary of mean Inflection Point and Convergence Point for the evolutionary strategies of Mutation and Crossover with Mutation.}
\label{tab:Summary}
\end{table*}

The trajectories of these experiments and the quantitative features of the punctuated equilibria depend sensitively on the random number generator that dictate the initial phenotypes of the bots and their mutations. Therefore, to establish any significant quantitative difference between the two evolutionary algorithms, a statistical study is performed. The experiments with Mutation, and Crossover with Mutation are each repeated 100 times with different random number seeds. This results in an ensemble of trajectories for each approach.

One may be naively tempted to consider the average of $T$ at each generation over the 100 trajectories for each ensemble. However, since the inflection happens at a different point in each experiment, such averaging would result in a soft falling curve and would thus lose information on where the inflections occur. To avoid this, we fit each of the 100 trajectories with the logistic function, extract the $g_0$ and $c$, and plot their distributions for comparison between the two evolutionary strategies.

Fig.~\ref{fig:inflection_point_mutation} and~\ref{fig:inflection_point_mutation_and_crossover} show the distributions of the Inflection Points in the 100 experiment ensembles for the inheritance algorithms of Mutation, and Crossover with Mutation, respectively. They are both fitted with Gaussians to extract the means and standard deviations of these distributions. We note that while Mutation inflects at $512 \pm 68$ generations, Crossover with Mutation inflects significantly earlier at $300 \pm 9$ generations. Thus, one may say Crossover with Mutation results in 40\% faster learning than just Mutation.

Fig.~\ref{fig:c_convergencePoint_mutation} and~\ref{fig:c_convergencePoint_MutationCrossover} show the distributions of the Convergence Points for Mutation, and Crossover with Mutation, respectively. While the distribution for Mutation may be fitted to a simple Gaussian with mean at $699 \pm 5$ time-steps, the distribution for Crossover with Mutation is clearly bi-modal. We fit the latter with the sum of two Gaussians and find that their means are at $759 \pm 25$ and $1967 \pm 31$ time-steps, respectively. This and the lack of a bi-modal distribution in Fig.~\ref{fig:inflection_point_mutation_and_crossover} imply that in a fair fraction of cases, Crossover with Mutation converge to a less-than-optimal solution though it starts the learning process faster. By comparing areas under the two peaks of the bi-modal distribution, we find that fraction to be 29\%. We summarize these results in Table~\ref{tab:Summary}.

\section{Conclusions}
\label{sec:Conclusions}

Spiking neural networks are the third generation of neural networks. They allow for encoding information in the temporal sequence of spikes and thus offer higher computational capacity. Further, the sparseness of spikes make them energy efficient and thus appropriate for neuromorphic applications. However, training them requires novel methods. In this paper, we have demonstrated a multi-agent ER based framework inspired by evolutionary rules and competitive intelligence to train SNNs for performing a task efficiently. Two evolutionary inheritance algorithms, Mutation and Crossover with Mutation, are demonstrated and their respective performances are compared over statistical ensembles. We find that Crossover with Mutation promotes 40\% faster learning in the SNN than mere Mutation with a statistically significant margin. We also note that Crossover with Mutation results in 29\% of experiments converging to a less-than-optimal solution.

Future directions of this work may lead to the integration of evolutionary approaches with in-lifetime learning models like reinforcement learning.

\pagebreak
\bibliography{SpikingEvolution}
\end{document}